\documentclass{article}
\usepackage{ijcai24}

\usepackage{times}
\usepackage{soul}
\usepackage{url}
\usepackage[hidelinks]{hyperref}
\usepackage[utf8]{inputenc}
\usepackage[small]{caption}
\usepackage{graphicx}
\usepackage{amsmath}
\usepackage{amsthm}
\usepackage{booktabs}
\usepackage{algorithm}
\usepackage{algorithmic}
\usepackage[switch]{lineno}

\usepackage{caption}

\usepackage{multirow}
\usepackage{subcaption}
\usepackage{booktabs}
\usepackage{pifont}
\usepackage{svg}
\usepackage{enumitem}
\usepackage{amssymb}
\usepackage{mathtools}
\usepackage{array}
\usepackage{tablefootnote}
\usepackage{dblfloatfix}

\newcommand{\tabnum}[2]{#1{\tiny (#2)}}
\newcommand{\besttabnum}[2]{\textbf{#1}{\tiny (#2)}}

\newcommand{\set}[1]{\mathcal{#1}}
\renewcommand{\vec}[1]{\boldsymbol{#1}}

\newcommand{\GED}{\operatorname{GED}}
\newcommand{\G}{{G}}
\newcommand{\V}{\mathcal{V}}
\newcommand{\E}{\mathcal{E}}
\newcommand{\model}{EPIC}
\newcommand{\etal}[1]{#1 \textit{et al}.}

\title{EPIC: Graph Augmentation with Edit Path Interpolation via Learnable Cost}

\author{
Jaeseung Heo$^{1}$
\and
Seungbeom Lee$^{1}$\and
Sungsoo Ahn$^{1,2}$\And
Dongwoo Kim$^{1,2}$\\
\affiliations
$^1${Graduate School of Artificial Intelligence, POSTECH, South Korea}\\
$^2${Department of Computer Science \& Engineering, POSTECH, South Korea}\\
\emails
\{jsheo12304, slee2020, sungsoo.ahn, dongwookim\}@postech.ac.kr
}

\begin{document}

\maketitle

\begin{abstract}
Data augmentation plays a critical role in improving model performance across various domains, but it becomes challenging with graph data due to their complex and irregular structure. 
To address this issue, we propose EPIC (\textbf{E}dit \textbf{P}ath \textbf{I}nterpolation via learnable \textbf{C}ost), a novel interpolation-based method for augmenting graph datasets.
To interpolate between two graphs lying in an irregular domain, EPIC leverages the concept of graph edit distance, constructing an edit path that represents the transformation process between two graphs via edit operations.
Moreover, our method introduces a context-sensitive cost model that accounts for the importance of specific edit operations formulated through a learning framework. This allows for a more nuanced transformation process, where the edit distance is not merely count-based but reflects meaningful graph attributes. 
With randomly sampled graphs from the edit path, we enrich the training set to enhance the generalization capability of classification models.
Experimental evaluations across several benchmark datasets demonstrate that our approach outperforms existing augmentation techniques in many tasks. 
\end{abstract}

\section{Introduction}

Graph data has become increasingly important in various domains, such as social networks, bioinformatics, and recommendation systems~\cite{hu2020open,Morris+2020,leskovec2007dynamics}. Despite the increasing importance, the limited size and diversity of existing graph datasets often limit the performance of graph-based models. One way to overcome this limitation is to augment the existing dataset, a technique that has found success in the other domains~\cite{shorten2019survey,iwana2021empirical}.

An interpolation between points is a recognized method for data augmentation, yet this approach does not seamlessly translate to graphs, due to the irregular structures inherent to them. Prior solutions have either adopted a continuous relaxation of graph structures into regular domains~\cite{han2022g} or employed transplantation of graph components~\cite{park2022graph}, thereby circumventing the difficulties associated with graph interpolation.

We propose a novel interpolation method for graphs based on the concept of graph edit distance~\cite{ullmann1976algorithm}. The graph edit distance, a widely-accepted metric, quantifies graph similarity by determining the minimum number of necessary edit operations, such as insertions, deletions, or substitutions of nodes and edges. From the graph edit distance between two graphs, we construct a graph edit path representing the transformation process from one graph to the other through the edit operations. The intermediate graphs along this path can be seen as an interpolation between two graphs and are used to augment the training set.

However, the conventional edit distance with operation counts may overlook the significance of individual operations, such as those that affect functional groups in molecular graphs~\cite{durant2002reoptimization,zhang2020motif}. In response to this, we propose a context-sensitive cost model built through an edit distance learning framework that maximizes the distance between graphs with different labels while minimizing those with the same label. This enables us to learn a cost model that better reflects the underlying graph data characteristics.

By combining the graph edit distance metric with the learned cost model, we present a novel graph dataset augmentation method named Edit Path Interpolation via learnable Cost (\model). Through rigorous experimental evaluations on various graph classification tasks, we demonstrate the superiority of our method in comparison to existing techniques. Additional experiments with noisy labels show the robustness of our approach against the others.

\section{Related Work}
\subsection{Graph Edit Distance}

The graph edit distance is a metric that quantifies the dissimilarity between two graphs~\cite{ullmann1976algorithm}. It measures the minimum number of edit operations required to transform one graph into another. These edit operations include node and edge insertion, deletion, and substitution. The computational complexity of obtaining graph edit distance is known to NP-complete~\cite{abu2015exact}.
A number of works addressing the problem of the high computational cost have been proposed. 
\etal{Cross}~\shortcite{cross1997inexact} casts the optimization process into a Bayesian framework with a genetic search algorithm.
\etal{Myers}~\shortcite{myers2000bayesian} adopts the Levenshtein distance to model the probability distribution for structural errors in the graph-matching. In Justice and Hero~\shortcite{justice2006binary}, a binary linear programming formulation of the graph edit distance for unweighted, undirected graphs is proposed.
\etal{Fischer}~\shortcite{fischer2015approximation} proposes an approximated graph edit distance based on Hausdorff matching. 
\etal{Wang}~\shortcite{wang2021combinatorial} adopts the deep learning method to efficiently prune the search tree in computation.

Recently, deep neural network-based graph edit distance learning methods have been proposed. In contrast to the traditional fixed cost distance, these methods learn the cost of individual edit operations.
One standard approach to learning costs using neural networks involves obtaining embeddings from node and edge attributes, which are then used to compute the edit distance in a supervised manner~\cite{cortes2019learning}. However, these approaches require ground truth node correspondences provided by an oracle between two graphs, which are inapplicable in various situations. \etal{Riba}~\shortcite{riba2021learning} learns the cost using node features extracted by graph neural networks and optimize the cost by approximating graph edit distance to Hausdorff distance. The Hausdorff distance is an effective method for approximating the graph edit distance in quadratic time, but it is not suitable for edit path construction since it allows one-to-many substitution between nodes. 

\subsection{Graph Augmentation}
Augmentation methods for graph-structured data aim to improve the generalization ability of neural networks by creating diverse training samples. The most commonly used augmentation methods are based on random modifications of original data. For example, DropNode \cite{feng2020graph} and DropEdge \cite{rong2019dropedge} uniformly drop nodes and edges, respectively. A subgraph sampling \cite{you2020graph,wang2020graphcrop} and motif swap \cite{zhou2020data} perturb the subgraph of the original graph via subgraph matching. FLAG \cite{kong2022robust} adds adversarial perturbation to node features. These methods assign the same labels before and after the perturbation.

To overcome the simplicity of basic approaches, mixup-based augmentation has been proposed. Manifold Mixup \cite{wang2021mixup} interpolates embeddings from the last layer for two graphs and uses it as a graph representation of the augmented graph.
SubMix~\cite{yoo2022model} proposes a node split and merge algorithm to perturb original graphs and then mix random subgraphs of multiple graphs. 
S-Mixup~\cite{ling2023graph} uses an alignment matrix from the graph matching network~\cite{li2019graph} to mix node features. 
G-Mixup~\cite{han2022g} mixes graphons~\cite{airoldi2013stochastic} extracted from different classes of graphs. 
GREA~\cite{liu2022graph} proposes a rationalization identification algorithm that extracts a subgraph explaining a property of the graph best and uses the subgraph to curate augmented graphs.

\section{EPIC: Edit Path Interpolation via Learnable Cost}
In this section, we first describe a graph data augmentation method with a graph edit path. We then propose a method to learn a graph edit distance by learning the cost of individual edit operations.

 \begin{figure*}
    \centering
    \includegraphics[width=0.9\textwidth]{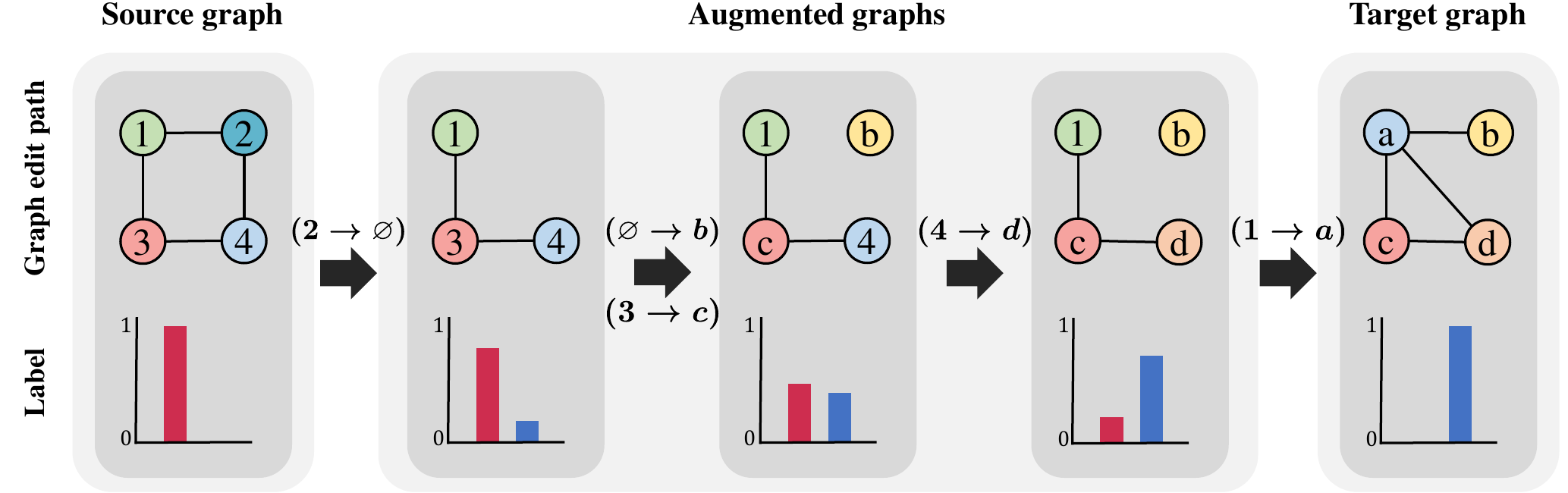}
    \caption{Illustration of graph edit path and their corresponding labels for augmentation.} 
    \label{fig:edit_path}
\end{figure*}

\subsection{Augmentation with Graph Edit Path}

\subsubsection{Construction of Graph Edit Path}
We consider a graph $\G = (\set{V}, \set{E})$ associated with node and edge attributes.
The graph edit distance is a metric that quantifies the dissimilarity between two graphs. It measures the minimum number of edit operations required to transform one graph into another or computes the total cost of edit operations if the cost of individual operations varies. These edit operations involve node and edge insertion, deletion, and attribute substitution.

Once the graph edit distance is computed, a graph edit path can be obtained by applying a series of edit operations from a source graph to reach a target graph. It represents the step-by-step transformation from one graph to another while minimizing the edit distance. 

The graph edit distance is generally invariant to the order of edit operations. However, there are certain dependencies between node and edge operations. The node deletion operation can only be performed after all the connected edges are deleted. The edge insertion operation can only be performed when the two target nodes are presented.

We only consider the node operations in order to reduce the computational complexity of graph edit distance. Specifically, when a new node is inserted into a graph, we perform edge insertion operations for all edges whose adjacent nodes are given after insertion. When a node is deleted, we perform edge deletion operations for all edges connected to the deleted node. When a node is substituted, we perform edge insertion, deletion, and substitution operations accordingly. The detailed algorithm is provided in Algorithm~\ref{alg:node-operation}. By doing so, we can construct the graph edit path whose length is equal to the number of node edit operations in the computation of the graph edit distance. \autoref{fig:edit_path} illustrates an example of a graph edit path between two graphs and possible augmentation. 

\begin{algorithm}[t!]
\caption{Applying node operation}
\label{alg:node-operation}
\begin{algorithmic}[1]
\REQUIRE A node operation $(u \rightarrow v)$, source graph $\G_S = (\V_S, \E_S)$, target graph $\G_T = (\V_T, \E_T)$, and the current graph $\G_C = (\V_C, \E_C)$
\ENSURE The updated graph $\G_{C+1} = (\V_{C+1}, \E_{C+1})$
\STATE $\G_{C+1} = \text{copy}(\G_C)$
\IF{$u=\varnothing$}
  \STATE Insert the node $v$ to $\V_{C+1}$.
\ELSIF{$v=\varnothing$}
  \STATE Delete the node $u$ from $\V_{C+1}$.
\ELSE
  \STATE Substitute the node $u$ with the node $v$ in $\V_{C+1}$.
\ENDIF
\FOR{each edge $(u,w)$ in $\E_C$}
  \IF{node $w \in \V_T$ and edge $(v,w)$ not in $\E_T$ }
    \STATE Remove the edge $(v,w)$ from $\E_{C+1}$.
  \ELSIF{node $w \in {\V_T}$ and edge $(v,w)$ in $\E_T$}
    \STATE Substitute edge attributes of $(v,w)$ in $\E_{C+1}$ with those of $(v,w)$ in $\E_T$ if needed.
  \ENDIF
\ENDFOR
\FOR{each edge $(v,x) \in \E_T $ adjacent to node $v$ in $\V_T$}
  \IF{node $x \in {\V_C}$ and edge $(u,x)$ not in $\E_C$}
    \STATE Insert the edge $(v,x)$ to $\E_{C+1}$.
  \ENDIF
\ENDFOR
\end{algorithmic}
\end{algorithm}

\subsubsection{Graph Augmentation and Label Assignment}
We use the graph edit path to construct an augmented graph. We randomly sample two graphs in the training set. The graph edit distance between the two graphs is computed, then the graph edit path is constructed with node operations in random order. The samples obtained from a graph edit path are used as augmented graphs. To prevent the case where the augmented graph is already presented in the training set, we add an additional node feature indicating whether the graph is from the augmentation set or not.

To assign a label to the augmented graph, we use the cost of edit operations from the augmented graph to the source and the target graphs. Let $(o_1, ..., o_n)$ be a sequence of edit operations applied to transforming source graph $\G_{S}$ to target graph $\G_{T}$, and $c(o)$ be the real-valued cost function of operation $o$, which is precisely defined in Subsection~\ref{subsec:learning}. With the corresponding one-hot classification label of the source graph $\boldsymbol{y}_S$ and the target graph $\boldsymbol{y}_T$, the label of the augmented graph $\bar{\boldsymbol{y}}$ obtained by applying the first $m$ operations $(o_1, ..., o_m)$ is computed as
\begin{align}
\label{label-equation}
    \bar{\boldsymbol{y}} = \frac{\sum_{i=m+1}^n c\left(o_i\right)}{\sum_{i=1}^n c\left(o_i\right)} \boldsymbol{y}_{S}+\frac{\sum_{i=1}^m c\left(o_i\right)}{\sum_{i=1}^n c\left(o_i\right)} \boldsymbol{y}_{T},
\end{align}
where $c(\cdot)$ measures the cost of an operation. The assigned label is inversely proportional to the operation cost to reach the source or target from the augmented graph.

\begin{figure*}[t!]
    \centering
    \includegraphics[width=0.87\textwidth]{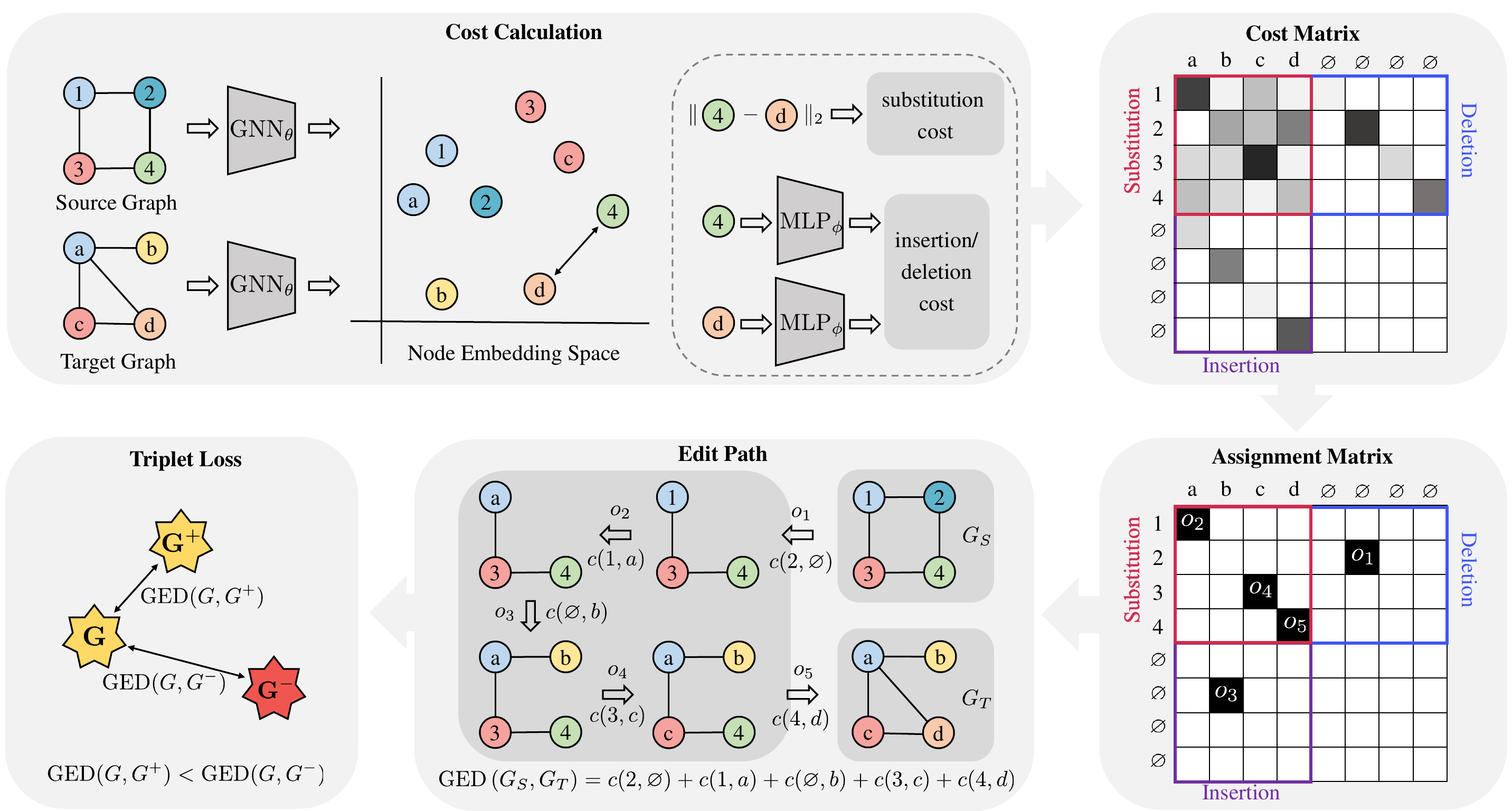}
    \caption{Overall illustration of graph edit distance learning. The assignment matrix can be obtained by either Hungarian or Sinkhorn-Knopp. For learning, we use Sinkhorn-Knopp, and for augmentation, we use Hungarian.}
    \label{fig:overall}
\end{figure*}

\subsection{Learning Costs of Edit Operations}
\label{subsec:learning}

The standard unit cost model \cite{ullmann1976algorithm,fischer2015approximation,riesen2009approximate} of the graph edit distance is incapable of measuring the importance of each operation as all operations are assigned the same cost, regardless of their significance or impact. However, the importance of edit operations would differ based on the context of the dataset.
For example, changes in a functional group of molecular graphs can lead to larger semantic perturbation than the other parts in a property prediction task~\cite{durant2002reoptimization,zhang2020motif}. Therefore, the edit operation leading to a large semantic modification should cost more than the others. To measure the importance of each operator in the computation of graph edit distance, we propose a learning algorithm for the operation cost based on a neural network model.

\subsubsection{Triplet Loss for Learning Distance}

A good cost function should be problem dependent. 
We use the triplet loss with known labels~\cite{schroff2015facenet} to learn the graph edit distance and the operation costs therein. 
We assume that the pair of graphs within the same class has a relatively shorter edit distance than those of different classes. 
Let $\GED(\G, \G')$ be the distance between graphs $\G$ and $\G'$.
We propose a triplet loss-based objective function to encode our intuition:
\begin{multline}
\label{eqn:objective}
\mathcal{L}(\G,\G^+, \G^-) \\
= \max\left(\GED(\G,\G^+) - \GED(\G,\G^-) + \gamma, 0\right),
\end{multline} 
where $\G^+$ is a positive example whose label is the same as $\G$, $\G^-$ is a negative example whose labels are different from $\G$, and $\gamma$ is a margin hyperparameter.

\subsubsection{Graph Edit Distance as Constrained Optimization}
The computational complexity of graph edit distance computation is NP-complete~\cite{abu2015exact}. Learning graph edit distance often requires relaxation to make the algorithm tractable~\cite{fischer2015approximation,riesen2009approximate}. To simplify the learning, we assume that the cost of node operation \emph{subsumes} the cost of dependent edit operations, similar to the construction of the edit path. With the simplified assumption, we only need to consider the following three cases when computing the edit distance between source graph $\G_S = (\V_S,\E_S)$ and target graph $\G_T = (\V_T,\E_T)$ with operation cost function $c: \V_{S} \cup \varnothing \times \V_{T} \cup \varnothing \rightarrow \mathbb{R}$, where $\varnothing$ represents an empty node:
\begin{itemize}
\setlength\itemsep{-0.5em}
\item Node $u$ in $\V_S$ is substituted by node $v$ in $\V_T$ with substitution cost $c(u, v)$.
\item Node $u$ in $\V_S$ is deleted with deletion cost $c(u, \varnothing)$.
\item Node $v$ in $\V_T$ is inserted with insertion cost $c(\varnothing, v)$.
\end{itemize}

We construct a cost matrix that encapsulates all required costs to compute the edit distance between two graphs. 
The cost matrix is constructed as
\begin{equation}
    \resizebox{0.9\columnwidth}{!}{
        $C=\left[\begin{array}{cccccc}
        c\left(u_1, v_1\right) & \cdots & c\left(u_1, v_m\right) & c\left(u_1, \varnothing\right) & \cdots & \infty \\
        \vdots & \ddots & \vdots & \vdots & \ddots & \vdots \\
        c\left(u_n, v_1\right) & \cdots & c\left(u_n, v_m\right) & \infty & \cdots & c\left(u_n, \varnothing\right) 
         \\
        c\left(\varnothing, v_1\right) & \cdots & \infty & \infty & \cdots & \infty \\
        \vdots & \ddots & \vdots & \vdots & \ddots & \vdots \\
        \infty & \cdots & c\left(\varnothing, v_m\right) & \infty & \cdots & \infty
        \end{array}\right]$
    }
\end{equation}
where $n$ and $m$ are the number of nodes in $\G_S$ and $\G_T$, respectively.

With the cost matrix, the problem of computing the graph edit distance can be reduced to solving the assignment problem.
Since the node in the source graph can only be substituted or deleted, only one operation can be performed in each of the first $n$ rows.  Similarly, since the node in the target graph can only be substituted or inserted, only one operation in each of the first $m$ columns can be performed in edit distance computation.
A binary assignment matrix $X$, whose size is the same as the cost matrix, is introduced to indicate which operation is performed in edit distance computation. With the assignment matrix, the computation of graph edit distance can be formulated as a constrained optimization problem
\begin{align}
\label{eqn:ged}
\operatorname{GED}(\G_S, \G_T) = &\min_{X}\sum_{i=1}^{n+m} \sum_{j=1}^{n+m}C_{ij} X_{ij} \notag \\[-10pt]
\text{s.t. } \sum_{j=1}^{n+m} X_{ij}=&1, \quad 1 \leq i \leq n \notag \\[-10pt]
\sum_{i=1}^{n+m} X_{ij}=1, \quad 1 \leq j &\leq m, \quad X_{ij} \in \{0,1\}.
\end{align}

\begin{table*}[t!]

\begin{center}


\renewcommand{\arraystretch}{1.2}
\resizebox{\linewidth}{!}{
\begin{tabular}{clcccccccccr}

\midrule[1.0pt]
 &{\small Method} & {\small NCI1} & {\small BZR} & {\small COX2} &{\small M}utagen. & {\small IMDB-B} & {\small IMDB-M}& {\small PROTEINS} & {\small ENZYMES} & Rank 
\\ 
\midrule[0.5pt]
\multirow{11}{*}{\rotatebox[origin=c]{90}{GIN}} & Vanilla 
& \tabnum{81.68}{0.8}
& \tabnum{87.07}{2.7} 
& \tabnum{83.40}{2.7} 
& \tabnum{81.57}{0.7} 
& \tabnum{72.90}{0.6}
& \tabnum{48.40}{1.2} 
& \tabnum{67.80}{2.2} 
& \tabnum{46.33}{2.3} 
& 5.4
\\
\cmidrule{2-11}
& DropEdge~\cite{rong2019dropedge}
&\tabnum{74.61}{0.7} 
&\tabnum{85.85}{0.4} 
&\tabnum{80.43}{2.7} 
&\tabnum{79.13}{1.1} 
&\tabnum{71.20}{1.6}
&\tabnum{47.20}{2.0} 
&\tabnum{68.97}{1.6} 
&\tabnum{39.00}{3.1} 
& 8.6
\\ 
 & DropNode~\cite{feng2020graph} 
&\tabnum{73.05}{1.8} 
& \tabnum{85.61}{3.0} 
&\tabnum{78.51}{1.6} 
& \tabnum{77.91}{1.0}
& \tabnum{72.90}{3.4}
& \tabnum{46.60}{2.0} 
& \tabnum{67.53}{3.3} 
& \tabnum{38.00}{3.6} 
& 9.4
\\
 & Subgraph~\cite{you2020graph} 
& \tabnum{74.90}{2.6} 
& \tabnum{79.76}{3.3} 
& \tabnum{81.49}{4.9} 
& \tabnum{69.25}{1.8}
& \tabnum{71.90}{3.1}
& \tabnum{46.13}{3.0} 
& \tabnum{63.32}{3.5} 
& \tabnum{34.50}{8.4} 
& 10.3
\\
 & FLAG~\cite{kong2022robust}
&\underline{\tabnum{81.85}{0.8}} 
& \underline{\tabnum{88.29}{1.6}}
& \tabnum{84.26}{2.6}
& \tabnum{81.98}{0.6}
& \tabnum{72.80}{0.3}
& \underline{\tabnum{48.60}{1.4}}
& \tabnum{68.34}{2.9} 
& \tabnum{46.17}{2.5} 
&\underline{4.1}
\\
\cmidrule{2-11}
 & SubMix~\cite{yoo2022model}
&\tabnum{81.31}{0.6}
& \tabnum{87.56}{3.5}
& \tabnum{83.62}{2.3}
& \tabnum{80.25}{1.1} 
& \tabnum{72.90}{0.9}
& \underline{\tabnum{48.60}{1.0}}
& \tabnum{70.76}{2.2}
& \tabnum{44.33}{4.8} 
& 4.8
\\
 & M-Mix~\cite{wang2021mixup}
&\tabnum{81.00}{0.4} 
& \tabnum{85.37}{3.6} 
& \besttabnum{84.47}{1.0}
& \tabnum{81.68}{0.9}
& \besttabnum{73.20}{0.6}
& \tabnum{47.93}{1.8} 
& \tabnum{68.70}{1.1} 
& \tabnum{46.67}{1.7}
&4.9
\\
 & G-Mix~\cite{han2022g}
& \tabnum{80.92}{1.9} 
& \tabnum{88.05}{1.6}
& \tabnum{82.98}{0.8} 
& \tabnum{81.66}{0.6}
& \tabnum{73.00}{0.9}
& \tabnum{48.13}{1.2} 
& \tabnum{69.87}{2.4} 
& \tabnum{44.83}{1.4} 
&5.1
\\

 & GREA~\cite{liu2022graph}
& \tabnum{74.63}{0.5} 
& \tabnum{85.85}{2.2}
& \tabnum{81.91}{2.4} 
& \tabnum{79.98}{0.4}
& \tabnum{72.40}{1.0}
& \tabnum{46.27}{0.7} 
& \besttabnum{73.36}{1.4} 
& \underline{\tabnum{46.83}{2.1}} 
&6.6
\\
& S-Mix~\cite{ling2023graph}
& \tabnum{81.48}{1.4} 
& \tabnum{85.85}{2.5}
& \tabnum{83.83}{0.5} 
& \underline{\tabnum{82.07}{1.1}}
& \tabnum{73.00}{2.0}
& \tabnum{48.47}{0.6} 
& \tabnum{69.42}{1.6} 
& \tabnum{44.33}{6.3} 
& 4.4
\\ 
\cmidrule{2-11}
 & \model
 & {\besttabnum{82.16}{0.7}}
 & \besttabnum{88.78}{3.3}
 & \besttabnum{84.47}{1.9}
 & \besttabnum{82.30}{0.5}
 & \besttabnum{73.20}{0.4}
 & \besttabnum{49.07}{1.2} 
 & \underline{\tabnum{71.21}{1.4}}
 & \besttabnum{47.50}{1.8} 
 &\textbf{1.1}
\\
\midrule[1.0pt]
\multirow{11}{*}{\rotatebox[origin=c]{90}{GCN}} 
& Vanilla 
&\tabnum{81.07}{0.7}
& \tabnum{85.85}{2.8}
& \tabnum{84.26}{2.2} 
& \tabnum{81.70}{1.1}
& \tabnum{70.60}{1.3}
& \besttabnum{48.27}{1.5} 
& \tabnum{64.13}{1.6} 
& \tabnum{44.67}{5.7} 
&5.1
\\
\cmidrule{2-11}
& DropEdge~\cite{rong2019dropedge}
&\tabnum{74.24}{1.0} 
& \tabnum{82.93}{4.3} 
&\tabnum{83.40}{1.2} 
&\tabnum{80.37}{0.6} 
& \tabnum{70.50}{1.8}
& \tabnum{45.67}{1.5} 
& \tabnum{64.75}{3.4} 
& \tabnum{39.33}{1.8} 
&8.0
\\ 
& DropNode~\cite{feng2020graph} 
&\tabnum{73.78}{1.8} 
& \tabnum{80.73}{2.4} 
&\tabnum{79.15}{1.8} 
& \tabnum{78.32}{1.6}
& \tabnum{69.40}{3.0}
& \tabnum{39.80}{3.7} 
& \tabnum{68.43}{1.8}
& \tabnum{36.17}{3.6} 
&9.1
\\
 & Subgraph~\cite{you2020graph} 
& \tabnum{70.11}{5.5} 
& \tabnum{77.80}{1.0} 
& \tabnum{78.09}{3.2} 
& \tabnum{79.47}{1.3}
& \tabnum{59.00}{6.6}
& \tabnum{39.07}{2.1} 
& \tabnum{66.28}{2.5} 
& \tabnum{28.33}{8.2} 
&10.5
\\
 & FLAG~\cite{kong2022robust}
& \tabnum{82.02}{0.6} 
& \underline{\tabnum{88.05}{2.3}} 
& \underline{\tabnum{85.32}{1.4}} 
& \tabnum{81.01}{0.8}
& \underline{\tabnum{71.00}{2.0}}
& \tabnum{47.33}{2.7} 
& \tabnum{67.62}{1.2} 
& \besttabnum{47.50}{4.4} 
&\underline{3.1}
\\
\cmidrule{2-11}
& SubMix~\cite{yoo2022model}
&\tabnum{81.99}{0.6} 
& \tabnum{86.34}{2.0} 
& \tabnum{84.68}{3.7} 
& \tabnum{80.99}{0.6} 
& \tabnum{70.30}{1.4}
& \tabnum{46.47}{2.5} 
& \tabnum{67.80}{2.0} 
& \tabnum{43.67}{4.9} 
&5.1
\\
& M-Mix~\cite{wang2021mixup}
&\tabnum{81.41}{0.5}
& \tabnum{84.15}{2.3}
& \tabnum{83.83}{2.1}
& \underline{\tabnum{81.96}{0.6}}
& \tabnum{69.40}{1.1}
& \tabnum{46.40}{2.7} 
& \tabnum{66.28}{1.5} 
& \tabnum{44.00}{5.3} 
&6.1
\\
& G-Mix~\cite{han2022g}
&\underline{\tabnum{82.04}{1.8}}
& \tabnum{87.32}{2.4}
& \tabnum{84.89}{1.4}
& \tabnum{80.32}{0.7}
& \tabnum{69.90}{1.8}
& \tabnum{45.87}{3.0} 
& \tabnum{68.07}{1.2}
& \underline{\tabnum{46.17}{4.3}}
&4.8
\\
 & GREA~\cite{liu2022graph}
& \tabnum{75.31}{0.4} 
& \tabnum{84.39}{0.9}
& \tabnum{80.43}{0.5} 
& \tabnum{80.48}{0.5}
& \tabnum{65.50}{0.0}
& \tabnum{45.73}{1.4} 
& \besttabnum{73.18}{0.7} 
& \tabnum{38.00}{2.2} 
&7.4
\\
& S-Mix~\cite{ling2023graph}
& \tabnum{81.65}{0.7} 
& \tabnum{87.56}{2.8}
& \tabnum{83.40}{1.4} 
& \tabnum{81.63}{0.4}
& \tabnum{69.60}{0.8}
& \tabnum{46.20}{0.8} 
& \tabnum{68.52}{0.8}
& \tabnum{45.00}{3.2} 
&4.8
\\
\cmidrule{2-11}
& \model
&\besttabnum{82.14}{0.8} 
& \besttabnum{88.78}{2.3} 
& \besttabnum{85.53}{1.6}
& \besttabnum{82.44}{0.7}
& \besttabnum{71.70}{1.0}
& \underline{\tabnum{47.93}{1.3}}
& \underline{\tabnum{69.06}{1.0}}
& \tabnum{45.00}{3.0}
&\textbf{1.5}
\\
\midrule[1.0pt]
\end{tabular}
}
\end{center}

\caption{Classification accuracy of TUDataset. We report the average and standard deviation (in brackets) over five seeds. We mark the best and the second-best performances in \textbf{bold} and \underline{underline}, respectively. The rank column shows the average rank of model performance across all datasets.}
\label{tab:Result of TUDataset}

\vspace{-1em}
\end{table*}

\subsubsection{Design Cost Function with Neural Networks}

We introduce the graph neural network framework to parameterize the cost function $c$. Specifically, we use the embedding distances between two nodes as the substitution cost. Let $h_{u}$ and $h_{v}$ be the output embedding of node $u \in \G_S$ and $v \in \G_T$ from a graph neural network. We use the distance between two embeddings as a substitution cost, i.e., $c_\theta(u, v) = ||h_{u} - h_{v}||_2$, where $\theta$ is the parameter of the graph neural network. The embeddings of the graph neural network encode the neighborhood structure of the target node. If the embeddings of two nodes are similar, then the two nodes are likely to play a similar role in the graph. Hence, the substitution cost measures the structural similarity between two nodes.
For the insertion and deletion operations, we additionally introduce a multi-layer perceptron, i.e., $c_{\theta,\phi}(u, \varnothing) = \operatorname{MLP}_\phi(h_u)$ and $c_{\theta,\phi}(\varnothing, v) = \operatorname{MLP}_\phi(h_v)$, where $\phi$ denotes the parameters of $\operatorname{MLP}$. The $\operatorname{MLP}$ computes the cost of insertion and deletion using the embedding of a node. We use the same network for both insertion and deletion.
The graph neural networks encode the local structure of nodes into node embeddings. Consequently, by considering the costs of node operations, we effectively encapsulate the information regarding the neighborhood edges as well.


\subsubsection{Optimization with a Differentiable Assignment Matrix}

To learn the graph edit distance, we need to minimize the loss in \autoref{eqn:objective} w.r.t $\theta$ and $\phi$. However, this optimization involves a non-differentiable optimization problem w.r.t the assignment matrix $X$ in \autoref{eqn:ged}.
The Hungarian algorithm~\cite{kuhn1955hungarian} can be used to find the optimal assignment $X$ for each iteration of the stochastic gradient descent step. However, the Hungarian algorithm is non-differentiable and has a computational complexity of $O(n^3)$, making it difficult to employ during gradient-based optimization.
We instead employ the Sinkhorn-Knopp algorithm~\cite{sinkhorn1967concerning} to address this issue to obtain a differentiable assignment matrix. The Sinkhorn-Knopp algorithm transforms a non-negative matrix into a doubly stochastic matrix to approximate the Hungarian algorithm. 
Specifically, Sinkhorn-Knopp iteratively updates a soft assignment matrix $\tilde{X}$ via two intermediate variables $\vec{u}$ and $\vec{v}$. Once $\vec{u}$ and $\vec{v}$ are initialized as a vector of ones, i.e., $\vec{u}^{(0)}= [1, \cdots, 1]^\top$, at $k$-th iteration of Sinkhorn-Knopp approximates the assignment matrix via 
\begin{align}
\label{sinkhorn-knop}
\vec{u}^{(k)} = \frac{X^{(k-1)}\mathbf{1}}{K \vec{v}^{(k-1)}}, \quad \vec{v}^{(k)} &= \frac{X^{{(k-1)}\top}\mathbf{1}}{K^\top \vec{u}^{(k-1)}}, \notag \\ 
\tilde{X}^{(k)} = \operatorname{diag}(\vec{u}^{(k)})  K  &\operatorname{diag}(\vec{v}^{(k)}),
\end{align}
where each entry of matrix $K$ is parameterized by the cost matrix and a regularizer parameter $\delta$ as $K_{ij} = \exp({-C_{ij}/\delta})$, and $\mathbf{1}$ is a vector of ones. $\delta$ is a regularization term controlling the sharpness of the assignment matrix.

Note that the back-propagation algorithm needs to optimize the entire iterative process of the Sinkhorn-Knopp approximation. In experiments, we set the number of maximum iterations to 10 to reduce the computational cost. After learning the cost function, to augment a pair of randomly selected graphs, we first compute the cost matrix and then create a graph edit path using the optimal assignment from the Hungarian algorithm. \autoref{fig:overall} shows the overall illustration of our proposed approach.

\section{Experiments}
In this section, we first show the effect of \model{} in graph classification tasks over 11 datasets. We further evaluate the robustness of GNNs with our method against corrupted labels. We provide additional analysis of our model selection process.

\subsection{Effect of Augmentation for Graph Classification} 
\label{sec:main_exp}
\paragraph{Datasets.}
We used eight classification datasets: NCI1, BZR, COX2, Mutagenicity, IMDB-BINARY, IMDB-MULTI, PROTEINS, ENZYMES from TUDataset~\cite{Morris+2020} and three classification dataset: BBBP, BACE, HIV from MoleculeNet~\cite{wu2018moleculenet}. The datasets cover a wide range of tasks, including social networks, bioinformatics, and molecules. 

\paragraph{Baselines.}
For baseline augmentation models, we use two graph augmentation methods as baselines: one that modifies a single graph, DropEdge~\cite{rong2019dropedge}, DropNode~\cite{feng2020graph}, Subgraph~\cite{you2020graph} and FLAG~\cite{kong2022robust}, another that mixes information from two graphs, SubMix~\cite{yoo2022model}, Manifold-Mixup (M-Mixup)~\cite{wang2021mixup}, G-Mixup~\cite{han2022g}, GREA~\cite{liu2022graph}, and S-Mixup~\cite{ling2023graph}. We also report the performance of a vanilla model without augmentation.



\paragraph{Implementation details.} 
We first learn the cost of edit operations for each dataset. We use Adam optimizer~\cite{kingma2014adam} with a learning rate decay of 0.1 every 25 epochs. We train the cost function for 100 epochs on TUDataset. While we use the Sinkhorn-Knopp approximation with $k=10$ in \autoref{sinkhorn-knop} for training, the Hungarian algorithm is used for inference to obtain an optimal assignment given costs. We perform graph classification tasks with GIN~\cite{xu2018powerful} and GCN~\cite{kipf2016semi} as a backbone model for augmentation. When we train each backbone model, for a fair comparison, we use the same hyperparameters and architecture tuned from the vanilla model for our method and other baseline models. We follow the Open Graph Benchmark setting~\cite{hu2020open} for MoleculeNet dataset.
When training classification models, we compute the edit path between randomly paired graphs in each batch and use randomly chosen graphs from the edit path as augmentation. We use the validation set to choose the portion of augmented data points. 

\paragraph{Classification results.}
\autoref{tab:Result of TUDataset} shows the overall results of the TUDataset for graph classification tasks. 
Our augmentation method outperforms the other baselines on seven and five datasets with GIN and GCN backbones, respectively, and achieves the second-best performance on one and two datasets with GIN and GCN. 
\autoref{tab:results of moleculenet} shows the AUC-ROC scores with MoleculeNet datasets. We achieve the best performance on all datasets except for one second-best performance with GIN and GCN backbones over other baselines. The results from two benchmarks show our augmentation method consistently improves the generalization of GNNs.

\begin{table}[t]
\centering

\renewcommand{\arraystretch}{1.1}
\resizebox{0.95\linewidth}{!}{
\footnotesize
\begin{tabular}{cp{14mm}>{\centering\arraybackslash}m{13mm}>{\centering\arraybackslash}m{13mm}>{\centering\arraybackslash}m{13mm}r}

\midrule[1.0pt]
& Method & BBBP & BACE & HIV & Rank
\\ 
\midrule[0.5pt]
\multirow{11}{*}{\rotatebox[origin=c]{90}{GIN}} & Vanilla 
&\tabnum{65.90}{1.9}
&\tabnum{77.01}{2.7} 
&\tabnum{75.10}{2.7} 
&6.3
\\
\cmidrule{2-6}
& DropEdge
&\tabnum{65.32}{1.8} 
& \tabnum{74.50}{0.9}
&\tabnum{75.57}{0.6}
&8.0
\\ 
& DropNode
&\tabnum{64.32}{5.0} 
& \tabnum{76.37}{3.0} 
&\tabnum{75.36}{1.3} 
&8.3
\\
& Subgraph
&\tabnum{62.85}{5.7} 
&\tabnum{76.15}{4.3} 
&\tabnum{73.84}{0.9} 
&10.0
\\
& FLAG
&\tabnum{65.64}{0.8} 
&\tabnum{79.45}{4.5} 
&\tabnum{75.24}{1.9} 
&5.7
\\
\cmidrule{2-6}
& SubMix
&\tabnum{65.58}{5.0} 
&\tabnum{77.50}{5.5} 
&\underline{\tabnum{76.36}{1.3}}
&4.3
\\
& M-Mix
&\tabnum{64.48}{1.6} 
&\tabnum{75.30}{2.6} 
&\tabnum{75.55}{2.0}
&8.0
\\
& G-Mix
&\tabnum{64.33}{3.2} 
&\tabnum{78.54}{1.9}
&\tabnum{76.06}{0.7} 
&5.7
\\ 
& GREA
&\underline{\tabnum{68.57}{0.6}} 
&\underline{\tabnum{79.97}{1.2}} 
&\tabnum{76.18}{1.1} 
&\underline{2.3}
\\
& S-Mix
&\tabnum{65.89}{2.0} 
&\tabnum{75.04}{5.5} 
&\tabnum{75.63}{3.0} 
&6.3
\\
\cmidrule{2-6}
& \model{}
&\besttabnum{68.88}{2.1} 
&\besttabnum{81.00}{1.1}
&\besttabnum{76.38}{0.3} 
&\textbf{1.0}

\\
\midrule[1.0pt]

\multirow{11}{*}{\rotatebox[origin=c]{90}{GCN}}& Vanilla 
&\tabnum{66.08}{3.4} 
&\tabnum{76.35}{4.2} 
&\tabnum{75.45}{0.7}
& 7.0
\\
\cmidrule{2-6}
& DropEdge
&\tabnum{65.71}{2.5} 
&\tabnum{72.79}{7.9} 
&\tabnum{75.90}{1.6} 
&7.7
\\ 
& DropNode
&\underline{\tabnum{68.33}{3.0}}
&\tabnum{71.37}{5.0} 
&\tabnum{74.41}{0.9}
&7.7
\\
& Subgraph
&\tabnum{63.09}{0.6} 
&\tabnum{76.39}{1.3} 
&\tabnum{73.75}{1.7} 
&9.3
\\
& FLAG
&\tabnum{68.14}{3.5} 
&\tabnum{77.87}{1.9}
&\tabnum{76.74}{3.2}
&\underline{3.3}
\\
\cmidrule{2-6}
& SubMix
&\tabnum{67.68}{1.2}
&\tabnum{75.19}{5.6}
&\tabnum{75.61}{2.1} 
&6.3
\\
& M-Mix
&\tabnum{67.38}{2.0}
& \besttabnum{79.67}{0.9}
& \tabnum{75.23}{1.2} 
&5.3
\\
& G-Mix
&\tabnum{65.10}{2.3} 
&\tabnum{77.54}{5.3}
&\tabnum{75.99}{0.9}
&6.0
\\ 
& GREA
&\tabnum{68.23}{1.9} 
&\tabnum{77.29}{4.0} 
&\underline{\tabnum{76.85}{1.7}} 
&\underline{3.3}
\\
& S-Mix
&\tabnum{65.65}{2.7} 
&\tabnum{74.04}{2.6} 
&\tabnum{75.32}{3.8} 
&8.7
\\
\cmidrule{2-6}
& \model{}
&\besttabnum{68.52}{1.7} 
&\underline{\tabnum{78.05}{1.3}}
&\besttabnum{77.30}{3.1}
&\textbf{1.3}

\\
\midrule[1.0pt]
\end{tabular}
}
\caption{Classification AUC-ROC of MoleculeNet}
\label{tab:results of moleculenet}
\end{table}

\subsection{Robustness Analysis}
\etal{Verma}~\shortcite{verma2019manifold} theoretically demonstrates that mixup-based augmentation enhances the robustness of deep neural networks against noisy labels. We also conduct a similar study.
Following the settings used in \etal{Han}~\shortcite{han2022g} and \etal{Ling}~\shortcite{ling2023graph}, we randomly corrupt labels in the training set of IMDB-BINARY, IMDB-MULTI, and Mutagenicity datasets and test the model performance with the uncorrupted test set. We run the experiments with three different proportions of noise: $\{0.2, 0.4, 0.6\}$ based on GIN. Except for the noise, we use the same setting used in Subsection~\ref{sec:main_exp}. \autoref{tab:Result of robustness} shows the classification accuracy with different proportions of noisy labels. \model{} outperforms the other baseline models, except for one case, showing the robustness of our augmentation under the noisy environment.

\begin{table*}[!t]

\begin{center}

\renewcommand{\arraystretch}{1.1}
\resizebox{\linewidth}{!}{
\begin{tabular}{lccccccccc}

\midrule[1.0pt]
&\multicolumn{3}{c}{IMDB-BINARY} & \multicolumn{3}{c}{IMDB-MULTI} & \multicolumn{3}{c}{Mutagenicity} \\
\cmidrule(lr){2-4}\cmidrule(lr){5-7}\cmidrule(lr){8-10}
Method & 20\% & 40\% & 60\% & 20\% & 40\% & 60\% & 20\% & 40\% & 60\%
\\ 
\midrule[0.5pt]
Vanilla 
&\tabnum{72.50}{1.2}
&\tabnum{66.10}{1.6} 
&\tabnum{44.30}{6.6} 
&\tabnum{48.00}{1.2}
&\tabnum{46.07}{1.1} 
&\tabnum{42.33}{3.7} 
&\tabnum{75.30}{0.7}
&\tabnum{68.52}{1.4} 
&\tabnum{46.38}{3.1} 
\\
DropEdge~\cite{rong2019dropedge}
&\tabnum{70.30}{4.1} 
&\tabnum{66.20}{5.1} 
&\tabnum{48.30}{4.0}  
&\tabnum{44.80}{2.4}
&\tabnum{42.60}{3.1} 
&\tabnum{37.93}{2.2} 
&\tabnum{75.47}{0.9}
&\tabnum{67.71}{1.5} 
&\tabnum{45.13}{4.1} 
\\ 
DropNode~\cite{feng2020graph} 
&\tabnum{70.90}{1.0} 
&\tabnum{68.80}{3.4} 
&\tabnum{52.40}{2.3}
&\tabnum{45.60}{1.9}
&\tabnum{43.00}{2.5} 
&\tabnum{39.53}{1.9} 
&\tabnum{72.98}{1.5}
&\tabnum{66.26}{1.5} 
&\tabnum{45.89}{3.1} 
\\
SubMix~\cite{yoo2022model}
&\tabnum{72.20}{1.0}
&\tabnum{67.70}{2.0}
&\tabnum{43.10}{9.4}
&\underline{\tabnum{48.67}{1.2}}
&\tabnum{45.53}{1.3} 
&\tabnum{40.60}{5.2} 
&\tabnum{73.00}{1.5}
&\tabnum{66.86}{2.6} 
&\besttabnum{59.03}{2.5} 
\\
ManifoldMix~\cite{verma2019manifold}
&\tabnum{71.90}{1.1} 
&\tabnum{69.90}{3.8} 
&\tabnum{49.70}{6.5}
&\tabnum{48.07}{1.3}
&\underline{\tabnum{46.67}{1.1}}
&\underline{\tabnum{43.40}{3.3}}
&\tabnum{75.49}{0.9}
&\tabnum{67.43}{3.6} 
&\tabnum{45.91}{2.7} 
\\
G-Mix~\cite{han2022g}
&\tabnum{71.60}{1.9} 
&\tabnum{70.00}{4.3}
&\tabnum{44.60}{10.0} 
&\tabnum{47.33}{0.8}
&\tabnum{44.80}{2.1} 
&\tabnum{40.67}{1.7} 
&\tabnum{76.06}{1.5}
&\underline{\tabnum{69.25}{0.8}}
&\tabnum{45.27}{4.1} 
\\
GREA~\cite{liu2022graph}
& \tabnum{71.80}{1.8} 
& \tabnum{67.50}{5.8}
& \tabnum{49.20}{1.8} 
& \tabnum{45.53}{1.5}
& \tabnum{44.13}{3.7}
& \tabnum{39.73}{1.2} 
& \underline{\tabnum{76.60}{1.8}} 
& \tabnum{68.63}{1.6} 
& \tabnum{49.64}{7.3} 
\\
S-Mix~\cite{ling2023graph}
&\underline{\tabnum{72.60}{1.0}}
&\underline{\tabnum{70.30}{3.6}}
&\underline{\tabnum{52.50}{7.4}}
&\tabnum{48.40}{0.5}
&\tabnum{46.60}{1.5}
&\tabnum{42.07}{4.3} 
&\tabnum{75.56}{0.5} 
&\tabnum{68.22}{3.8} 
&\underline{\tabnum{58.50}{5.3}}
\\
\cmidrule{1-10}
\model 
&\besttabnum{72.90}{1.0} 
&\besttabnum{71.30}{3.4}
&\besttabnum{52.60}{10.3} 
&\besttabnum{49.00}{0.9}
&\besttabnum{47.13}{1.7} 
&\besttabnum{44.40}{3.7} 
&\besttabnum{76.78}{1.2}
&\besttabnum{69.48}{1.9} 
&\tabnum{53.90}{5.6}
\\
\midrule[1.0pt]
\end{tabular}
}
\end{center}

\caption{Robustness analysis on IMDB-BINARY, IMDB-MULTI, and Mutagenicity datasets.}

\label{tab:Result of robustness}

\end{table*}

\begin{figure}[t!]
    \centering
    \includegraphics[width=\linewidth]{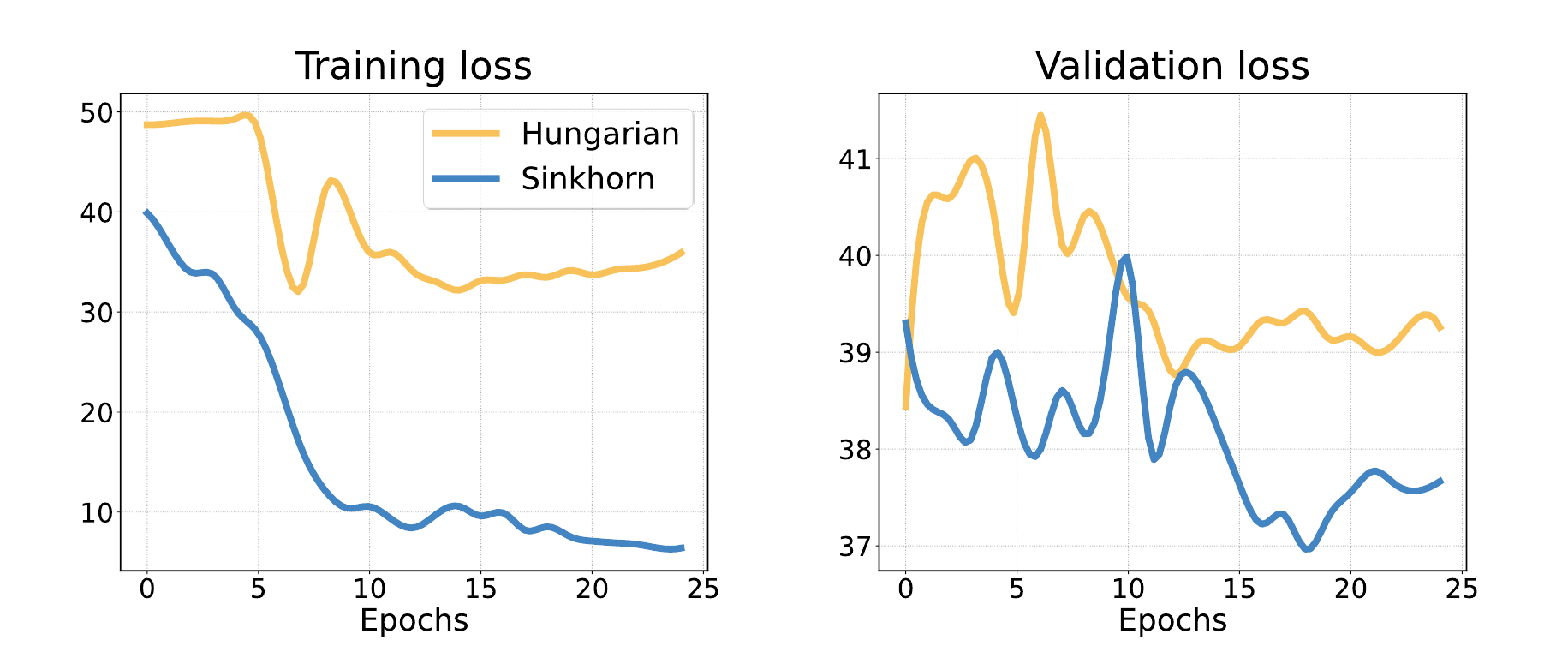}  
    \caption{Training and validation loss curves on HIV dataset with Hungarian and Sinkhorn-Knopp algorithms in the training process.}
    \label{sinkhorn_effectiveness}
\end{figure}
\subsection{Ablation Studies}
\label{sec:main_ablation}
In this subsection, we show the result of ablation studies on each component in \model. Further ablation studies are shown in \autoref{sec:ablations}.

\paragraph{Cost function variations.}
We test the effectiveness of the learnable cost function against other variations of the cost function. We use two variations of the cost function: unit cost, which measures the number of edit operations, and feature-distance cost, which measures the distance between two input node features, adopted from \etal{Ling}~\shortcite{ling2023graph}. The result in \autoref{tbl:cost_abal} shows that the learnable cost outperforms the other cost functions across all datasets. The results empirically verify our claim that the good cost function should be problem-dependent and can be learned from the dataset. We also classify the graphs in the test dataset based on their distance to the closest graph in each class. If the cost is learned properly, the distance from a test graph to the graph in the same class should be close to each other. \autoref{tbl:cost_abal_distance} shows the result of distance-based classification. In most cases, \model{} outperforms the other fixed-cost methods. Results for all TUDataset can be found in \autoref{appendix:additional_results}.

\begin{table}[t!]
\renewcommand{\arraystretch}{1.0}

\begin{subtable}[h]{\linewidth}
\centering

\resizebox{\linewidth}{!}{
\begin{tabular}{lcccc}

\midrule[1.0pt]
Method & NCI1~$\uparrow$ & Mutagen.$\uparrow$ & IMDB-B$\uparrow$ & IMDB-M~$\uparrow$
\\ 
\midrule[0.5pt]
Unit cost
& \tabnum{81.56}{1.3} 
& \tabnum{81.73}{0.2} 
& \tabnum{71.90}{0.9}
& \tabnum{48.53}{1.0}

\\
Feature-distance
& \tabnum{81.07}{0.9}
& \tabnum{81.68}{0.8}
& \tabnum{72.20}{1.2}
& \tabnum{48.53}{1.0}

\\ 
\model{}
& \besttabnum{81.70}{1.4}
& \besttabnum{82.30}{0.5}
& \besttabnum{73.20}{0.4}
& \besttabnum{49.07}{1.2}

\\
\midrule[1.0pt]
\end{tabular}
}
\vspace{-0.5em}
\caption{\label{tbl:cost_abal} Classification accuracy with augmentation}
\end{subtable}

\begin{subtable}[h]{\linewidth}
\centering

\resizebox{\linewidth}{!}{
\begin{tabular}{lcccc}
\midrule[1.0pt]
 Method & NCI1~$\uparrow$ & Mutagen.$\uparrow$ & IMDB-B$\uparrow$ & IMDB-M~$\uparrow$
\\ 
\midrule[0.5pt]
Unit cost
& \tabnum{60.49}{0.4} 
& \tabnum{56.69}{2.0} 
& \tabnum{55.50}{2.8}
& \tabnum{36.33}{2.2} 

\\
Feature-distance
& \tabnum{60.70}{0.8} 
& \tabnum{55.51}{1.8} 
& \tabnum{56.00}{2.8}
& \tabnum{36.20}{1.6} 

\\ 
\model{}
& \besttabnum{72.69}{0.3}
& \besttabnum{75.33}{1.0}
& \besttabnum{67.70}{1.8}
& \besttabnum{39.33}{2.8}
\\
\midrule[1.0pt]
\end{tabular}
}

\vspace{-0.5em}
\caption{\label{tbl:cost_abal_distance} Distance-based classification accuracy}
\end{subtable}

\caption{Comparisons between three different cost functions.}
\label{tab:differnt_cost}

\end{table}

\paragraph{Hungarian vs Sinkhorn-Knopp.}
We investigate the impact of a differentiable assignment matrix with the Sinkhorn-Knopp algorithm against a fixed assignment matrix from the Hungarian algorithm in the process of training. 
In \autoref{sinkhorn_effectiveness}, we present the training and validation curve on the HIV dataset. In general, the Sinkhorn-Knopp algorithm shows a more stable learning process than the Hungarian algorithm. The training loss with Hungarian has not been stabilized after 20 epochs in HIV datasets. Moreover, the validation loss of Sinkhorn-Knopp is consistently lower than Hungarian after certain iterations. 
We conjecture that the non-smooth loss surface of Hungarian makes the gradient descent work hard, and eventually, the model fails to reach a good local minimum, whereas the smooth loss surface of Sinkhorn-Knopp results in a better performance despite being an approximation of Hungarian.

\begin{figure}[t!]
\label{example of lollipo graph}
\begin{subfigure}{\linewidth}
  \centering
  \includegraphics[width=.8\linewidth]{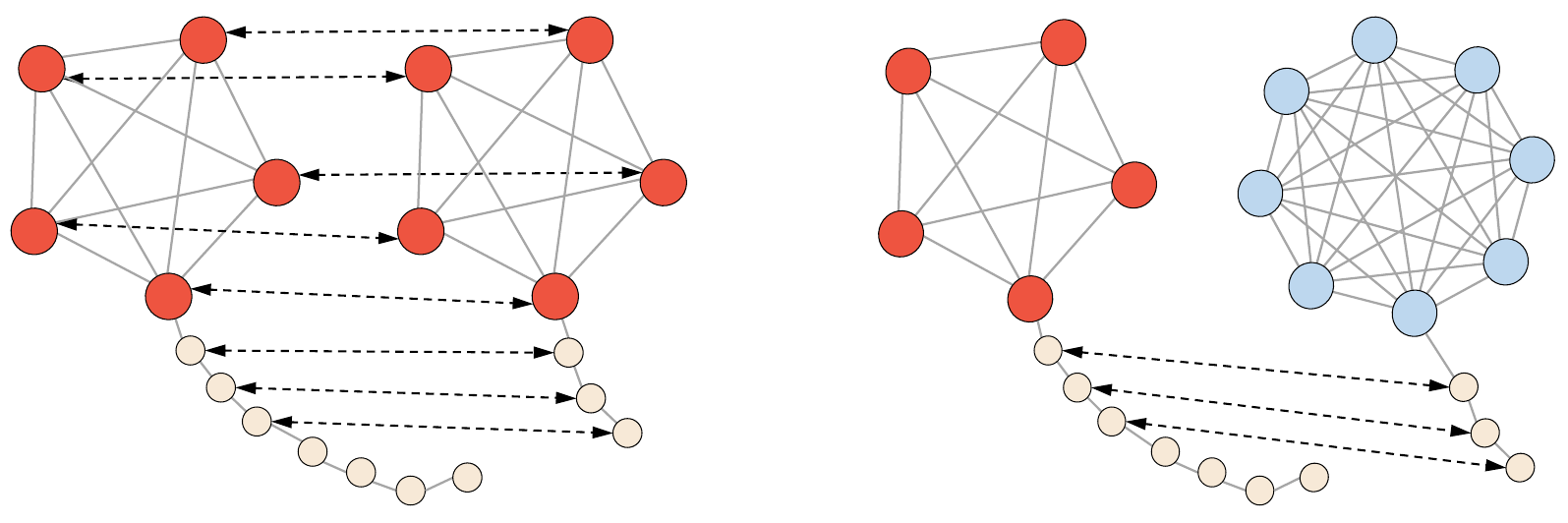}  
  \caption{Nodes with substitution operations}
  \label{fig:postive-pair}
\end{subfigure}

\medskip

\begin{subfigure}{\linewidth}
  \centering
  \includegraphics[width=.8\linewidth]{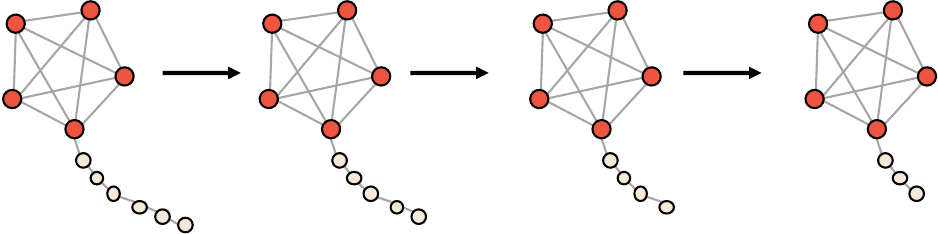}  
  \caption{Sampled edit path}
  \label{fig:postive-path}
\end{subfigure}

\caption{Examples of learned edit distance from a lollipop dataset. The dashed lines in (a) represent substitution operations.}

\label{fig:correspondence-path}
\end{figure}

\subsection{Qualitative Analysis}
To examine how the results of the learned distance are reflected in the augmented graph, we conduct an experiment with a lollipop dataset whose structural properties are easily visualized.
The $(m,n)$-\emph{lollipop} graph consists of a head, a complete graph with $m$ nodes, and a tail, a chain structured $n$ nodes. The lollipop dataset consists of graphs with varying $m$ and $n$. The label of a graph is the size of the head, i.e., $m$.
\autoref{fig:correspondence-path} displays examples of the trained graph edit distances and the corresponding edit path. With a positive pair, the learned graph edit distance substitutes the head nodes from the source graph for those of the target graph. Eventually, it maintains the complete subgraph of the head along the edit path. Additionally, we present the assigned cost for substitution to each node in \autoref{appendix:further analysis on lollipop}, which consistently shows the effectiveness of the cost function for \model.

\section{Conclusion}

In this paper, we propose a novel approach for graph dataset augmentation based on the graph edit distance. 
Our method overcomes the limitations of linear interpolation techniques in the non-Euclidean domain and provides a tailored augmentation solution for graph data. Through extensive experiments on benchmark datasets, we have demonstrated the effectiveness of our approach in improving the performance and robustness of graph-based models.



\begin{table*}[!t]

\begin{center}



\renewcommand{\arraystretch}{1.1}
\resizebox{0.95\linewidth}{!}{
\begin{tabular}{clcccccccccc}

\midrule[1.0pt]
& & \multicolumn{10}{c}{$k$} \\
\cmidrule(lr){3-12}
 &{\small D}ataset & 1 & 2 & 3 & 4 & 5 & 6 & 7 & 8 & 9 & 10
\\ 
\midrule[0.5pt]
\multirow{5}{*}{\rotatebox[origin=c]{90}{GIN}} & {\small BZR} 
& \tabnum{87.56}{2.3}
& \tabnum{87.32}{2.4}
& \besttabnum{89.51}{2.5}
& \tabnum{86.59}{2.6}
& \tabnum{88.05}{1.6}
& \tabnum{87.32}{1.8}
& \tabnum{87.32}{2.4}
& \tabnum{87.32}{2.0}
& \tabnum{88.05}{3.2}
& \tabnum{88.78}{3.3}
\\
& {\small COX2}
& \tabnum{82.98}{1.8}
& \tabnum{82.13}{3.6}
& \tabnum{83.19}{4.2}
& \tabnum{83.40}{2.7}
& \tabnum{82.55}{2.9}
& \besttabnum{84.68}{1.8}
& \tabnum{81.06}{4.4}
& \tabnum{84.47}{1.6}
& \tabnum{84.47}{1.2}
& \tabnum{84.47}{1.9}

\\
& {\small IMDB-B}
& \tabnum{72.30}{1.3}
& \tabnum{72.20}{0.8}
& \tabnum{72.80}{0.6}
& \tabnum{72.90}{0.5}
& \tabnum{72.60}{0.7}
& \tabnum{72.70}{0.9}
& \tabnum{72.70}{0.8}
& \tabnum{72.50}{0.7}
& \tabnum{72.60}{1.1}
& \besttabnum{73.20}{0.4}

\\
& {\small IMDB-M}
& \tabnum{48.40}{0.7}
& \tabnum{48.33}{0.2}
& \tabnum{48.67}{0.9}
& \tabnum{48.53}{0.7}
& \besttabnum{49.27}{0.6}
& \tabnum{48.67}{0.2}
& \tabnum{48.93}{0.9}
& \tabnum{49.13}{1.6}
& \tabnum{48.33}{0.8}
& \tabnum{49.07}{1.2}
\\
& {\small ENZYMES}
& \tabnum{45.00}{3.4}
& \tabnum{44.83}{3.2}
& \tabnum{46.50}{3.9}
& \tabnum{43.67}{4.7}
& \tabnum{46.00}{5.5}
& \tabnum{47.67}{2.2}
& \tabnum{45.17}{2.4}
& \besttabnum{49.33}{3.2}
& \tabnum{46.00}{3.7}
& \tabnum{47.50}{1.8}

\\
\midrule[0.5pt]
& {\small Rank}
& 7.4
& 8.6
& 3.8
& 6.8
& 4.6
& 3.6
& 6.2
& 3.8
& 5.0
& \textbf{2.2}

\\
\midrule[1.0pt]
\end{tabular}
}
\end{center}
\caption{Classification performance under varying values of $k$.}

\label{tab:sink_it}
\end{table*}
\begin{table*}[!t]

\begin{center}



\renewcommand{\arraystretch}{1.0}
\resizebox{0.8\linewidth}{!}{
\begin{tabular}{clcccccccccc}

\midrule[1.0pt]
 &{\small M}ethod & {\small NCI1} & {\small BZR} & {\small COX2} &{\small M}utagen. & {\small IMDB-B} & {\small IMDB-M}& {\small PROTEINS} & {\small ENZYMES}
\\ 
\midrule[0.5pt]
\multirow{2}{*}{\rotatebox[origin=c]{90}{GIN}} & BFS 
& \tabnum{81.75}{0.8}
& \besttabnum{88.78}{2.0}
& \besttabnum{84.68}{2.7}
& \besttabnum{82.51}{1.0}
& \tabnum{73.00}{0.5}
& \besttabnum{49.60}{0.8}
& \tabnum{70.94}{1.8}
& \tabnum{46.83}{2.5}
\\
 & Random
 & \besttabnum{82.16}{0.7}
 & \besttabnum{88.78}{3.3}
 & \tabnum{84.47}{1.9}
 & \tabnum{82.30}{0.5}
 & \besttabnum{73.20}{0.4}
 & \tabnum{49.07}{1.2} 
 & \besttabnum{71.21}{1.4}
 & \besttabnum{47.50}{1.8} 
\\
\midrule[1.0pt]
\end{tabular}
}
\end{center}
\caption{Comparison of randomly ordered augmentation and BFS-ordered augmentation. }

\label{tab:operation_order}
\end{table*}
\appendix

\section{Further Ablation Studies}

\label{sec:ablations}
In this section, we describe the ablation studies for model configuration. Specifically, we examine methods for determining the operation order in the edit path and the number of iterations in the Sinkhorn-Knopp algorithm.

\begin{figure}[t]
    \centering
    \includegraphics[width=0.9\linewidth]{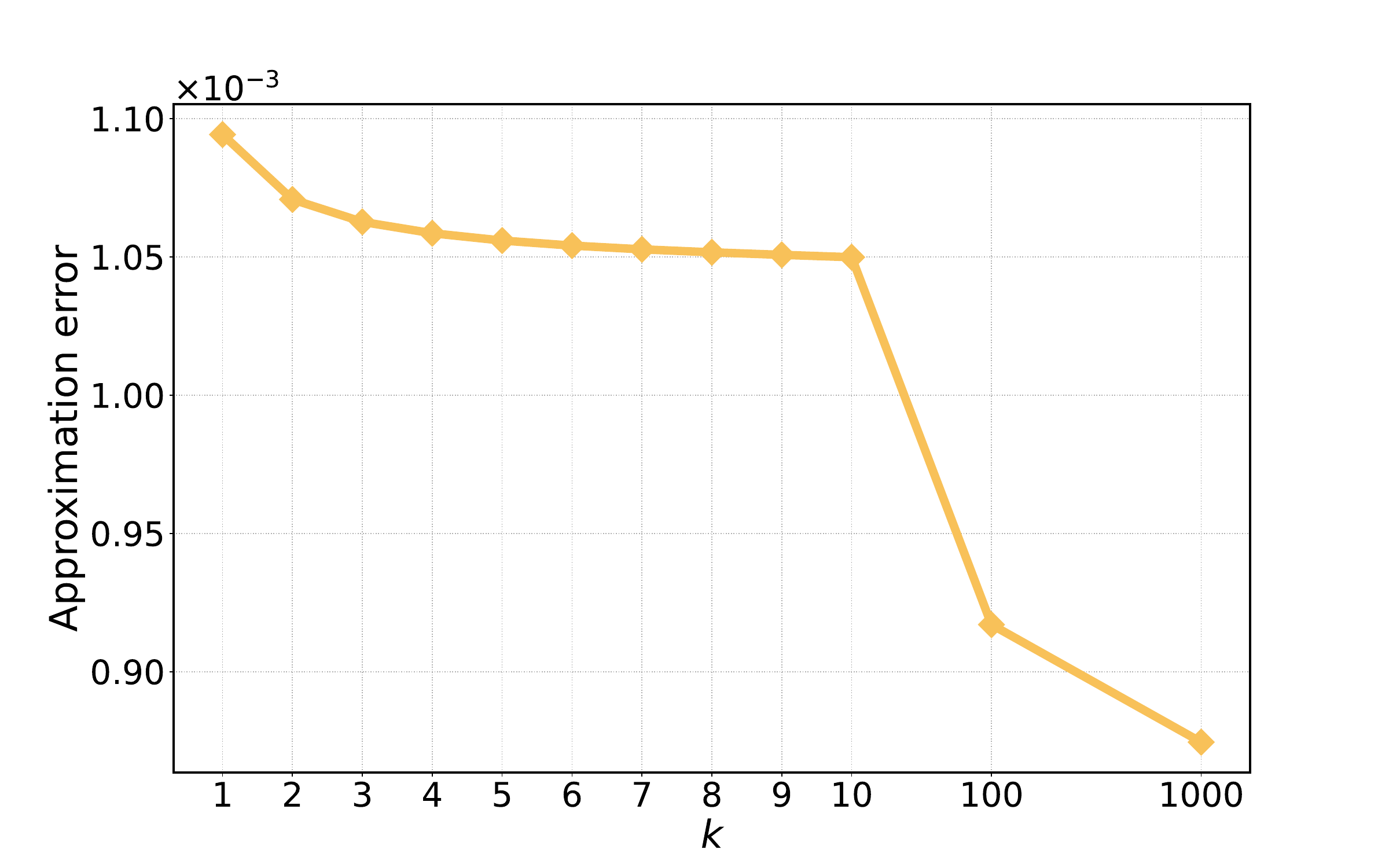}
    \caption{Difference between assignment matrix from Hungarian and Sinkhorn algorithm with varying $k$ in Equation~\ref{sinkhorn-knop}.}
    \label{fig:sinkhorn_error}
\end{figure}

\paragraph{Iterations of the Sinkhorn-Knopp algorithm.}
We adopt the Sinkhorn-Knopp algorithm to approximate the Hungarian algorithm to make the assignment matrix differentiable for optimization during training as described in Subsection~\ref{subsec:learning}.
To analyze the effect of the number of iterations of the Sinkhorn-Knopp algorithm, we evaluate the classification performance under varying values of $k$ in \autoref{sinkhorn-knop}. \autoref{tab:sink_it} presents the results of an ablation study.
We further experiment to observe how well the assignment matrix, $X_{S}$, derived from the Sinkhorn-Knopp algorithm approximates the assignment matrix, $X_{H}$, from the Hungarian algorithm, as the parameter $k$ varies. Firstly, we calculate both assignment matrices from our learned cost function on the test dataset and compute the Frobenius norm of the difference between the two matrices, $||X_{A}||_{F} = ||X_{S}-X_{H}||_{F}$, and then, measure an average of each Frobenius norm across the entire dataset. \autoref{fig:sinkhorn_error} shows the average of Frobenius norm with varying $k$. We observe that the error decreases as $k$ increases, suggesting that the Sinkhorn-Knopp algorithm's assignment matrix becomes more similar to the assignment matrix from the Hungarian algorithm. However, as $k$ increases, the computational cost required for training also increases, indicating a trade-off relationship.

\begin{table*}[!t]
\label{tab:appendix_differnt_cost}
\renewcommand{\arraystretch}{1.0}

\begin{subtable}[h]{\linewidth}
\centering
\resizebox{0.9\linewidth}{!}{
\begin{tabular}{lcccccccc}

\midrule[1.0pt]
Method & NCI1~$\uparrow$ & BZR$\uparrow$ & COX2$\uparrow$ & Mutagen.$\uparrow$ & PROTEINS$\uparrow$ & ENZYMES$\uparrow$ & IMDB-B$\uparrow$ & IMDB-M~$\uparrow$
\\ 
\midrule[0.5pt]
Unit cost
& \tabnum{81.56}{1.3} 
& \tabnum{85.37}{2.7} 
& \tabnum{84.04}{2.5} 
& \tabnum{81.73}{0.2} 
& \tabnum{70.13}{0.9} 
& \tabnum{45.33}{3.2} 
& \tabnum{71.90}{0.9}
& \tabnum{48.53}{1.0}

\\
Feature-distance
& \tabnum{81.07}{0.9}
& \tabnum{87.80}{2.3}
& \tabnum{82.34}{2.4}
& \tabnum{81.68}{0.8}
& \tabnum{71.12}{.12} 
& \tabnum{47.33}{1.8} 
& \tabnum{72.20}{1.2}
& \tabnum{48.53}{1.0}

\\ 
\model{}
& \besttabnum{81.70}{1.4}
& \besttabnum{88.78}{3.3}
& \besttabnum{84.47}{1.9}
& \besttabnum{82.30}{0.5}
& \besttabnum{71.21}{1.4}
& \besttabnum{47.50}{1.8} 
& \besttabnum{73.20}{0.4}
& \besttabnum{49.07}{1.2}

\\
\midrule[1.0pt]
\end{tabular}
}

\caption{\label{tbl:Appendix_cost_abal} Classification accuracy with augmentation}
\vspace{1em}

\end{subtable}

\begin{subtable}[h]{\linewidth}
\centering
\resizebox{0.9\linewidth}{!}{
\begin{tabular}{lcccccccc}
\midrule[1.0pt]
Method & NCI1~$\uparrow$ & BZR$\uparrow$ & COX2$\uparrow$ & Mutagen.$\uparrow$ & PROTEINS$\uparrow$ & ENZYMES$\uparrow$ & IMDB-B$\uparrow$ & IMDB-M~$\uparrow$
\\ 
\midrule[0.5pt]
Unit cost
& \tabnum{60.49}{0.4} 
& \tabnum{72.20}{3.7} 
& \tabnum{52.34}{3.8} 
& \tabnum{56.69}{2.0} 
& \tabnum{56.86}{1.5} 
& \besttabnum{25.83}{3.0} 
& \tabnum{55.50}{2.8}
& \tabnum{36.33}{2.2} 

\\
Feature-distance
& \tabnum{60.70}{0.8} 
& \tabnum{72.93}{4.0} 
& \tabnum{52.55}{3.2} 
& \tabnum{55.51}{1.8} 
& \besttabnum{57.31}{1.7} 
& \besttabnum{25.83}{2.9} 
& \tabnum{56.00}{2.8}
& \tabnum{36.20}{1.6} 

\\ 
\model{}
& \besttabnum{72.69}{0.3}
& \besttabnum{77.80}{2.9}
& \besttabnum{60.43}{3.5}
& \besttabnum{75.33}{1.0}
& \tabnum{56.95}{0.8} 
& \tabnum{23.67}{4.7} 
& \besttabnum{67.70}{1.8}
& \besttabnum{39.33}{2.8}
\\
\midrule[1.0pt]
\end{tabular}
}
\caption{\label{tbl:Appendix_cost_abal_distance} Distance-based classification accuracy}

\end{subtable}
\caption{Comparisons between three different cost functions. 
Three different cost functions: unit, feature-distance, and \model{}, are tested on TUDataset. GIN is used as a backbone.
}
\end{table*}

\paragraph{Operation ordering.}
When calculating the edit distance, the specific operation order does not affect the final result. To construct the edit path, we need to determine the operation order. In our analysis, we investigate the difference between two choices of operation orders. To maintain the local connectivity of the target graph within the edit path, we employ a breadth-first search (BFS) on the target graph, and node insertion/deletion/substitution is performed following the BFS order. The results in \autoref{tab:operation_order} compare randomly ordered augmentation with BFS-ordered augmentation on eight datasets. We find no significantly better ordering choices between these two methods. %
To minimize additional computational overhead, we utilize random ordering for the main result.

\section{Results on TUDataset Based on Cost Function Variations}
\label{appendix:additional_results}
In this subsection, we show the full result of classification described in \autoref{tab:differnt_cost}  based on different cost functions: the unit cost, which counts edit operations, and the feature-distance cost, derived from \etal{Ling}~\shortcite{ling2023graph}, which measures the distance between input node features. \autoref{tbl:Appendix_cost_abal} demonstrates that the learnable cost outperforms other cost functions on all datasets. The result empirically proves our claim that an effective cost function should be problem dependent and learned from the dataset. 

Also, if the cost is learned correctly, the graph with the shortest graph edit distance to a given graph should be in the same class. Under this premise, we classify the graphs in the test dataset based on their distance to the closest graph in each class. \autoref{tbl:Appendix_cost_abal_distance} shows the result of distance-based classification. In most cases, \model{} outperforms the other fixed-cost methods.

\begin{figure}[t!]
\centering
\begin{subfigure}{0.45\textwidth}
  \includegraphics[width=\linewidth]{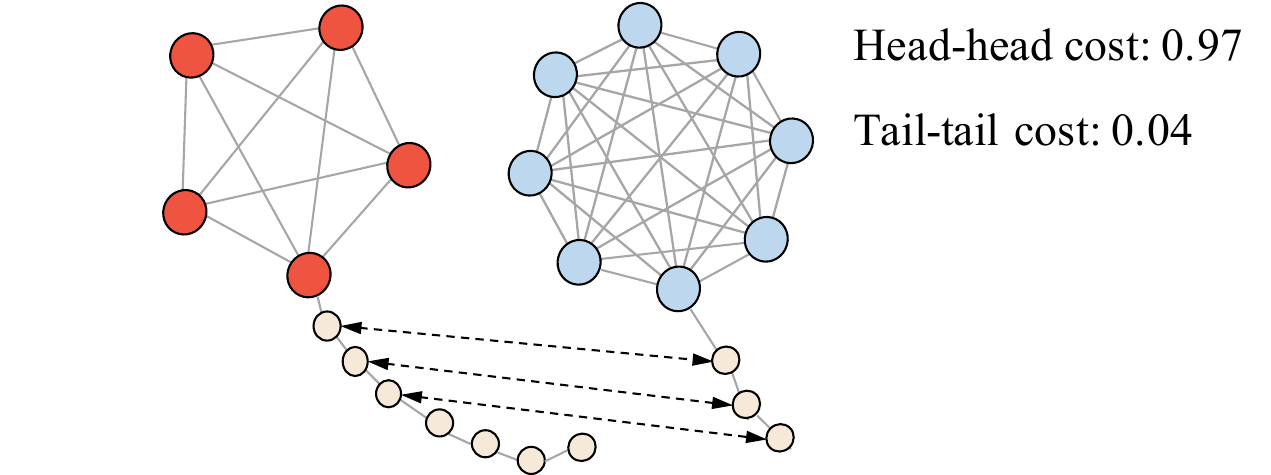}    
  \caption{Average substitution costs between negative pair}
  \label{appendix_cost_negative}
  \vspace{2em}

\end{subfigure}

\begin{subfigure}{0.45\textwidth}
  \includegraphics[width=\linewidth]{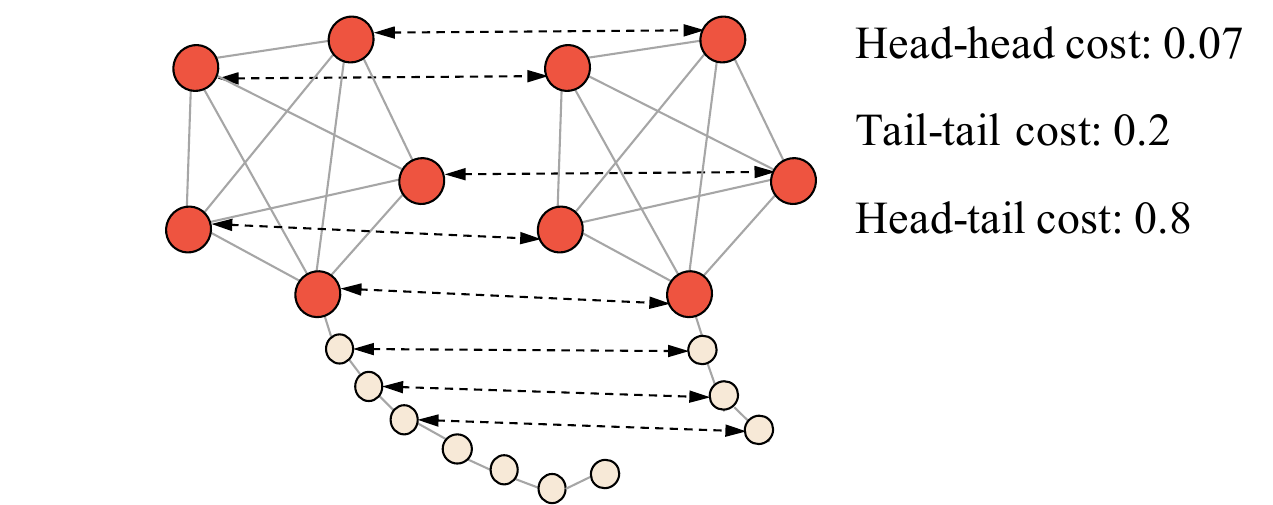}  
  \caption{Average substitution costs between positive pair}
  \label{appendix_cost_positive}
\end{subfigure}
\caption{Analysis of substitution cost on \emph{lollipop} graph. Normalized average substitution costs between source nodes and target nodes for \emph{negative} pair are shown in Figure~\ref{appendix_cost_negative} and \emph{positive} pair in Figure~\ref{appendix_cost_positive}. The costs are averaged across the entire dataset. }
  \vspace{2em}

\label{Appendix:cost analysis}
\end{figure}

\section{Further Analysis on \emph{Lollipop} Dataset}
\label{appendix:further analysis on lollipop}

To further show the effectiveness of the cost function for \model, we analyze the cost for substitution to each node on the \emph{lollipop} dataset. $(m,n)$-\emph{lollipop} graph consists of a head, a complete graph with $m$ nodes, and a tail, a chain structured $n$ nodes, with varying $m$ and $n$. We use the number of heads as the label of a graph, i.e., $m$. \autoref{Appendix:cost analysis} shows the head-to-head, head-to-tail, and tail-to-tail substitution costs. For instance, the head-to-head substitution cost measures the average substitution costs between the head nodes from the source graph with those of the target graph. 

With the \emph{negative pair}, the average node substitution cost for the head nodes is $0.97$, whereas the average for the tail nodes is $0.04$. These results align with our intuition that editing the head, closely related to the label, incurs a high cost. Thus, any modification to this part leads to substantial changes in the label. 
With the \emph{positive pair}, the head-to-head and tail-to-tail substitution costs are $0.07$ and $0.2$, respectively. Those values are much smaller than $0.8$ of head-tail costs. The lower cost for the head-head substitution in a positive pair resulted from the fact that it involves the exchange of nodes associated with the same label.
Similarly, the low cost for the tail-to-tail substitution aligns well with our intuition since the nodes are unrelated to the label. In contrast, the head-to-tail substitution is assigned a high cost because it substitutes a label-associated part with a non-associated part. This preserves the head part of every graph on the edit path between \emph{positive}-pair, ensuring that the augmented graph keeps the same label over the path.

\section*{Acknowledgements}
\addcontentsline{toc}{section}{Acknowledgements}
This work was supported by Institute for Information \& communications Technology Promotion (IITP) grant funded by the Korea government (MSIP) (No. RS-2019-II191906: Artificial Intelligence Graduate School Program (POSTECH)), Basic Science Research Program through the National Research Foundation of Korea(NRF) funded by the Ministry of Education (2022R1A6A1A03052954, 2022R1C1C1013366, RS-2024-00337955), and the Technology Innovation Program (No. 20014926, Development of BIT Convergent AI Architecture, Its Validation and Candidate Selection for COVID19 Antibody, Repositioning and Novel Synthetic Chemical Therapeutics) funded by the Ministry of Trans, Industry \& Energy (MOTIE, Korea).
\section*{Contribution Statement}
Jaeseung Heo and Seungbeom Lee contributed equally to the paper as the first authors.

\bibliographystyle{test}
\bibliography{ijcai24}


\end{document}